\documentclass[preprint,12pt]{elsarticle}

\usepackage{amssymb}
\usepackage{amsmath}
\usepackage{bbding}
\usepackage{subcaption}
\usepackage{multirow}
\usepackage{comment}

\journal{Journal of the Franklin Institute}

\begin{document}

\begin{frontmatter}



\title{Improving Facial Emotion Recognition through Dataset Merging and Balanced Training Strategies} 


\author{Serap Kırbız} 

\affiliation{organization = {MEF University,Department of Electrical and Electronics Engineering},
            addressline={Huzur Mah. Maslak Ayazağa Cad. No:4 Sarıyer}, 
            city={İstanbul},
            postcode={34396}, 
            country={Türkiye}}

\begin{abstract}
In this paper, a deep learning framework is proposed for automatic facial emotion  based on deep convolutional networks. In order to increase the generalization ability and the robustness of the method, the dataset size is increased by merging three publicly available facial emotion  datasets:  CK+, FER+ and KDEF. Despite the increase in dataset size, the minority classes still suffer from insufficient number of training samples, leading to data imbalance. The data imbalance problem is minimized by online and offline augmentation techniques and random weighted sampling. Experimental results demonstrate that the proposed method can recognize the seven basic emotions with $82\% $ accuracy. The results demonstrate the effectiveness of the proposed approach in tackling the challenges of data imbalance and improving classification performance in facial emotion recognition.    
\end{abstract}


\begin{highlights}
\item \textbf{Proposed an automatic facial emotion recognition (FER) method} using deep convolutional networks , combining data merging, online/offline augmentation, and random weighted sampling for improved classification performance.
\item \textbf{Merged three distinct facial emotion datasets (CK+, FER+ and KDEF)} to create a comprehensive training dataset, enabling more robust learning and better generalization.
\item \textbf{Face alignment and landmark detection} were applied using RetinaFace, significantly enhancing the quality of facial images and ensuring consistency across datasets.
\item \textbf{Implemented both offline and online data augmentation techniques}, including RandAugment, which boosted model generalization and classification accuracy, especially for minority emotion classes.
\item \textbf{Random weighted sampling} was applied to address class imbalance, improving precision, recall, and F1 score for underrepresented classes such as disgust and fear.
\end{highlights}

\begin{keyword}
facial emotion recognition \sep  convolutional neural networks \sep face alignment \sep data augmentation \sep facial landmarks \sep random weighted sampling \sep 


\end{keyword}

\end{frontmatter}



\section{Introduction}
\label{sec:intro}
In interpersonal communication, facial expressions are the most important signs of emotions. Emotion recognition is the process of detecting emotions, a task that humans can perform with a large variability depending on their abilities and experiences. In recent decades, automatic recognition of emotions has been a popular research area in machine learning, since it has practical use in many applications such as healthcare \cite{dhuheir2021emotion}, marketing \cite{ribeiro2017deep}, human-computer interaction \cite{abdat2011human}, and security \cite{an2023deep}. 

In automatic facial emotion recognition (FER) systems, the aim is to classify the features extracted from facial expressions into seven basic emotions: anger, fear, happiness, surprise, sad, disgust and neutral \cite{ekman1971, matsumoto1992more}. Traditional FER systems consist of two distinct tasks: feature extraction and classification of the extracted features into seven emotion categories. The first step extracts facial features using methods such as Local Binary Patterns \cite{shan2009facial}, Scale-Invariant Feature Transform (SIFT) \cite{shi2020improved}, and Non-negative Matrix Factorization \cite{zhou2016method}. The second step classifies the extracted features into seven basic emotions using a classifier such as Support Vector Machines (SVM) \cite{abdulrahman2015facial} and K-Nearest Neighbor (KNN) \cite{thakare2016facial}. The traditional methods work fine on simple datasets in which the intra-class variation is low. The limitation of traditional FER systems is that it is challenging to improve the performance of the system when feature extraction and classification are separate tasks. 

With the improvements obtained in computational power and deep learning methods, particularly in Convolutional Neural Network (CNN) architectures used in image classification applications, several models have been developed for deep-learning-based FER. CNNs \cite{kayaouglu2023cnn, ezerceli2022convolutional, borgalli2022deep, jumani2019facial} have significantly improved FER performance, as they can extract hierarchical features directly from raw pixel data. However, many FER datasets still have many limitations in terms of size, diversity, and imbalances in class distributions. 

In this work, we propose a deep learning based facial emotion recognition method to  improve FER accuracy using several strategies. Our main contributions are summarized as follows: 
\begin{itemize}
    \item We present a novel FER approach based on deep CNN architectures that combines data merging, online/offline augmentation, and random weighted sampling for improved classification performance.
    \item To address the challenges of data scarcity and imbalance in FER datasets, we propose creating a larger and more comprehensive training set by merging three publicly available datasets: Extended Cohn-Kanade (CK+) \cite{lucey2010extended}, Karolinska Directed Emotional Faces (KDEF) \cite{calvo2008facial}  and Facial Emotion Recognition (FER+)  \cite{barsoum2016training}.
  
    \item To enhance feature extraction and minimize the impact of variations in lighting and background, we employ RetinaFace \cite{deng2020retinaface}, a robust facial landmark detection method. RetinaFace helps isolate the most informative facial regions (such as eyes, eyebrows, and mouth) and minimizes variations in lightning and background, contributing to more accurate emotion classification.
    
    \item Despite the increase in data size, dataset imbalance problem continues to affect model performance. To increase robustness and variations in the dataset, we first apply on-the-fly data augmentation using the random augmentation method, RandAug \cite{cubuk2020randaugment}.  

    \item The aligned faces and faces obtained from cropped landmarks using RetinaFace can be viewed as new data generated by applying some transformations to the original images. Therefore, we have also considered adding these images to the existing dataset to create a larger dataset before training. This process can be viewed as an offline augmentation strategy that enhances both flexibility and efficiency.
          
    \item In imbalanced datasets, the majority classes often dominate the training process, causing the model to overfit to these classes. To address this issue and improve learning for minority classes, we propose using random weighted sampling \cite{shekelyan2022weighted}.
   
\end{itemize}

The rest of the paper is organized as follows: Section \ref{sec:related} reviews related studies on deep learning-based FER systems. Section \ref{sec:proposed} introduces the proposed method. Section \ref{sec:results} evaluates the performance of the proposed method, and Section \ref{sec:conclusion} summarizes the key findings and discusses potential future research directions.

\section{Related Work}
\label{sec:related}
In this section, the main steps in facial emotion recognition is explained: pre-processing, feature extraction through deep learning and classification of emotions.  
\subsection{Preprocessing}
\label{sec:preprocess}
In order to reduce the variations within the images in the dataset, preprocessing is applied before training the deep neural network for feature extraction. 

\subsubsection {Face Alignment}
\label{sec:alignment}
Automatic face localization is applied prior to many image analysis tasks such as age analysis \cite{kwon1999age}, facial expression recognition \cite{li2022}  and face recognition \cite{tolba2006face}. The aim of face detection is to identify the face within an image and eliminate background data. The Viola-Jones algorithm \cite{chaudhari2018face} has been widely used for face detection in many tasks due to its robustness and computational efficiency.  

In FER applications, face detection is the only required preprocessing that needs to be applied before feature learning. However, aligning faces using facial landmarks can greatly enhance FER performance. By localizing facial landmarks, face alignment determines the distinctive and precise parts of faces \cite{merget2018}, reducing the impact of background, lighting conditions, facial size, and face rotation.

Traditional face alignment methods rely on geometric and statistical models. Some of the most commonly used methods in the literature are Active Appearance Models (AAMs) \cite{cootes2001active}, Active Shape Models (ASMs) \cite{cootes1995active}, and Active Contour Models \cite{kass1988snakes}. These models require extraction of features manually and, they are computationally expensive.
Moreover, their performance is poor in the presence of variations in facial expressions, illumination, and head pose. 

Deep learning based approaches enable  end-to-end learning, enhancing face alignment accuracy by extracting hierarchical features directly from the data. Chen et al. \cite{chen2014joint} proposes to jointly detect and align faces with random forest on differences of pixel intensities. Inspired by \cite{chen2014joint}, faces and five facial landmarks are detected jointly using multitask cascaded convolutional networks (MTCCN) \cite{zhang2016joint}. However, it has not been proved whether MTCNN performs well with small face images. The method described in  \cite{he2017mask} incorporates an additional branch for object mask prediction, alongside the main branch, which performs bounding box classification and regression. The incorporated additional branch improves the detection performance. Similarly, Chaudhuri et al. \cite{chaudhuri2019joint} propose an end-to-end network to predict the bounding box locations and 3D morphable model (3DMM) simultaneously. This method achieves better face detection performance in-the-wild by combining 2D information from bounding boxes and 3D information  from 3DMM parameters. In \cite{deng2020retinaface}, a single-shot method, RetinaFace, is proposed to jointly learn various face localisation tasks, e.g. face detection, 3D face reconstruction and 2D face alignment. RetinaFace \cite{deng2020retinaface} accurately detects faces, performs 2D face alignment and 3D face recognition and yields faster speed.  

\subsubsection{Data Augmentation}
\label{sec:augmentation}
Deep neural networks need a large dataset of labeled data to perform good generalization capabilities for a specific task. However, the publicly available FER datasets do not have sufficient labeled training data. Data augmentation becomes vital if increasing the dataset size is not possible. Generally, the data augmentation is performed on-the-fly to avoid overfitting. The training samples are randomly cropped, shifted, scaled, flipped horizontally or rotated by a degree to increase the variations in the dataset. 

In addition to on-the-fly augmentation, offline augmentation is also commonly used. In offline augmentation, some transformations are applied offline prior to training, thus saved as additional data in the dataset. As a result, the dataset then includes multiple augmented versions of each image, enriching the training data and improving model robustness. 

Recently, automated augmentation techniques have been developed which automatically selects one of the several augmentation techniques on the fly, such as AutoAugment \cite{cubuk2019autoaugment}, RandAugment \cite{cubuk2020randaugment} and AugMix \cite{hendrycks2019augmix}. Augmix \cite{hendrycks2019augmix} combines geometric and photometric transformations into a single image using pixel-wise operations. AutoAugment \cite{cubuk2019autoaugment} uses reinforcement learning to select the optimal augmentation from several augmentation techniques, while RandAugment \cite{cubuk2020randaugment} applies one or more techniques randomly. Among these, RandAugment \cite{cubuk2020randaugment} increases the FER accuracy significantly \cite{kirbiz2024facial}. 

\subsection{Feature Extraction Using Deep Neural Networks}
The aim in feature extraction is to extract key facial expression features from facial images. The extracted features are given as input to the classifier to classify the features into one of the seven basic emotions. 

Traditional methods typically focus on geometric or textual properties of the face. Geometric features are extracted from the spatial relationships between facial landmarks such as eyes, eyebrows, nose, and mouth \cite{cootes2001active}. Even these methods are computationally efficient for small datasets, they are sensitive to variations in head pose and occlusions. On the other hand, texture-based features are extracted from variations in pixel intensities, with Local Binary Patterns (LBP) \cite{prakasa2016texture}  and Histogram of Gradients (HOG) \cite{jumani2019facial} being the most commonly used features. These methods are simple to implement and effective under small variations, but their performance degrades with more challenging databases where the intra-class variation is high. 

In recent years, deep learning methods, especially Convolutional Neural Networks (CNNs) \cite{ezerceli2022convolutional, kayaouglu2023cnn} have revolutionized FER performance. Unlike traditional methods, deep learning based methods extract facial features automatically in an end-to-end manner, capturing both low-level features like mouth and eyes, and high-level features, such as overall facial expressions simultaneously through the hierarchical layers. This improves robustness and generalization. However, CNNs require a large labeled dataset to avoid overfitting and their performance can decrease in the presence of some challenges such as gender, age, and cultural differences in facial expressions; imbalanced datasets; variations in lightning, pose, and image quality. To overcome the overfitting problem encountered with deep learning based methods, many studies benefit from pre-trained models that have been trained on large datasets for other tasks such as face recognition. These models are then fine-tuned on smaller size FER datasets to adapt the learned features specifically for the emotion recognition  task. Kaya et al. \cite{kaya2017video} showed that VGG-Face  trained on the large-scale ImageNet dataset and fine-tuned on the FER2013 dataset, improves the accuracy of FER. This approach allows the transfer of the existing knowledge from another task; thus converges faster and generalizes better.

Some CNN-based methods further enhance performance by focusing on the most relevant facial features using attention mechanisms \cite{wang2020region, minaee2021deep}. These  mechanisms assign greater weight to the most relevant regions of the face, such as the mouth, eyes, and eyebrows. 

While most research in FER focuses on still images, there are some works that combine facial features in consecutive frames. For instance, Zhang et al. proposed the Parallel Hierarchical Recurrent Neural Network (PHRNN) \cite{zhang2017facial} that can learn complex temporal and spatial features from facial structures in videos and improves FER accuracy. These temporal dependencies can also be used to recognize micro-expressions. Another method, Hierarchical Transformer Network (HTNet) \cite{wang2024htnet} uses a transformer layer that captures local temporal features and an aggregation layer that extracts local and global facial features, especially for micro-expression recognition in videos. A more recent approach,  the Four Player GroupGAN \cite{niu2022four}, has been proposed for recognizing weak facial expressions.

\subsection{Classification of Facial Expressions}
\label{sec:classification}

In the final stage, the learned features are classified into one of the seven basic emotions. 

In traditional methods, feature extraction and classification are applied independently. In deep learning based approaches, the learned features are classified using an additional layer in an end-to-end way. In CNN-based methods, softmax loss is used generally to minimize the cross-entropy between the ground-truth distribution and the estimated distribution of the classes \cite{li2022}.   

\section{Proposed Work}
\label{sec:proposed}
The proposed method first detects and align the faces in order to remove the redundant part of the facial images that do not convey any information about the emotion. The preprocessed images are then trained by deep CNN models using data augmentation and random weighted sampling.  

\subsection {Data Preprocessing}

RetinaFace \cite{deng2020retinaface} has been preferred for face detection since RetinaFace achieves high detection accuracy even in difficult conditions where MTCNN \cite{zhang2016joint} fails such as small faces, low lightning conditions etc. RetinaFace allows to detect faces with high accuracy and also enables to automatically discard the images that do not contain faces, herey improving the quality of the dataset. 

RetinaFace \cite{deng2020retinaface} is a single-shot multi-level face localization method that was trained on the WIDER FACE dataset \cite{yang2016wider}. The algorithm  detects faces and  outputs five facial landmarks: right eye, left eye, nose, left mouth corner and right mouth corner.  It is particularly effective in handling small and challenging faces due to its robustness and accuracy among various image sizes and orientations. The method uses Feature Pyramid Networks (FPNs) to detect multi-scale features and a high-resolution backbone network, ResNet50, to extract detailed feature maps which is crucial for detecting small faces. 

\begin{figure}[t]
\centering
\includegraphics[width=\linewidth]{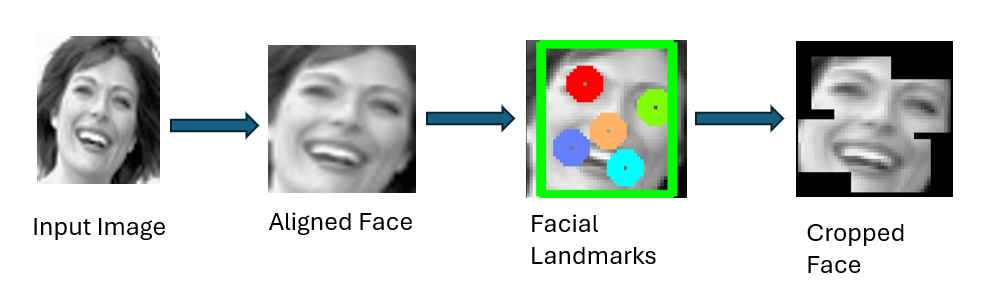}
\caption{The preprocessing steps applied to images for face alignment and landmark detection. }
\label{fig:preprocess}
\end{figure}

The preprocessing steps applied to input images before training are displayed in Figure \ref{fig:preprocess}. The input image is already a facial image. RetinaFace \cite{deng2020retinaface} detects faces and creates a bounding box around the detected faces, which is then cropped. The pixel coordinates of the five facial landmarks are detected in the next stage. In the proposed work, we use a rectangular box around each landmark and apply a mask to the facial image to remove the parts of the face which is not included in these regions. The cropped image is displayed in the last step of Figure \ref{fig:preprocess}.  

\begin{figure}
    \begin{subfigure}{0.58\textwidth}
    \includegraphics[width=\textwidth]{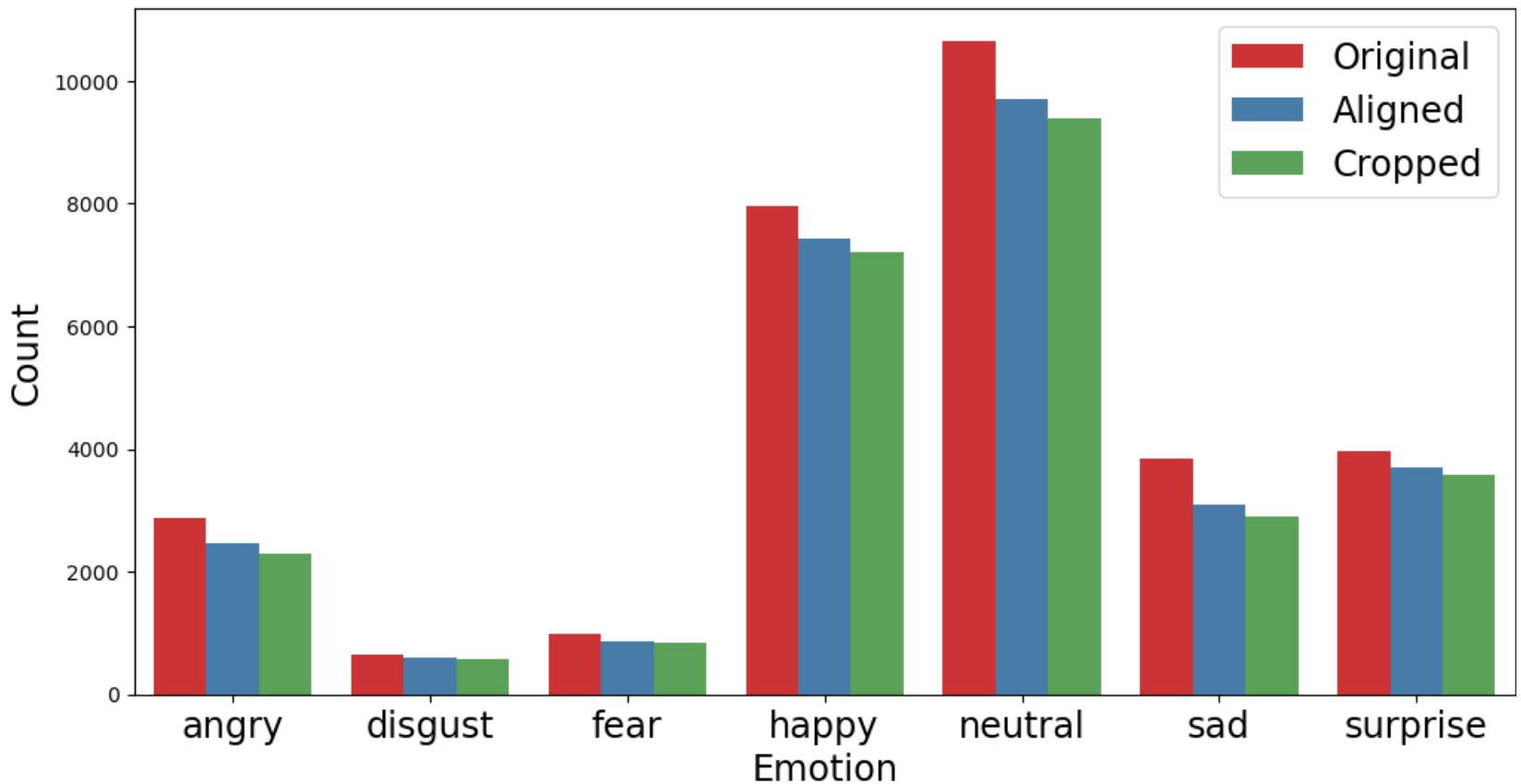}
    \caption{}
    \label{fig:data_distribution}
\end{subfigure}
\begin{subfigure}{0.42\textwidth}
    \includegraphics[width=\textwidth]{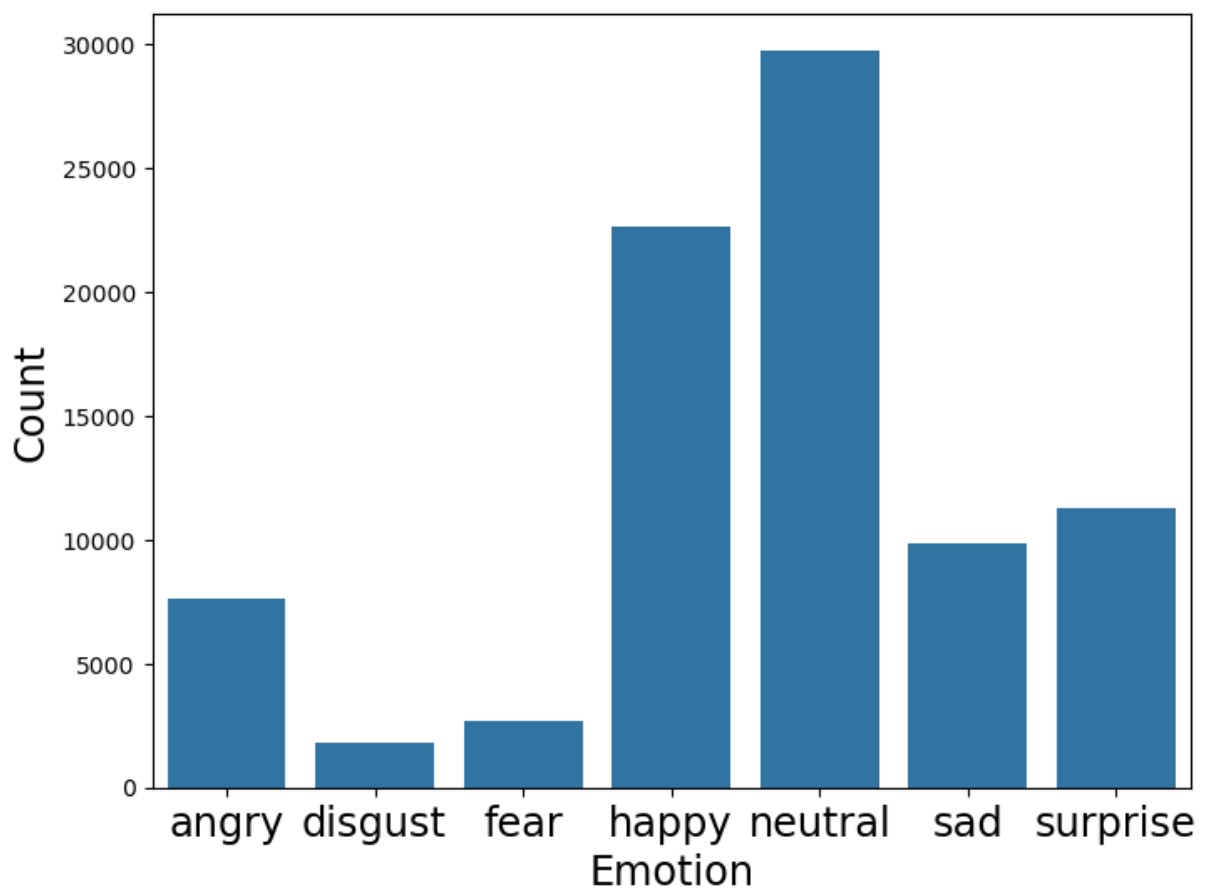}
    \caption{}
    \label{fig:merged_dataset}
\end{subfigure}
\caption{The distribution of the data: (a) for each class and for each stage of the data preprocess, (b) for the augmented merged dataset consisting of original, aligned, cropped images.}
\label{fig:dataset}
\end{figure}

The distribution of the dataset has been displayed in Figure \ref{fig:dataset}. Figure \ref{fig:data_distribution} displays the distribution of the original dataset, aligned and cropped dataset. The original dataset  contains the images from three datasets: FER+, CK+ and KDEF. The aligned dataset is obtained from the detected faces in the original dataset using RetinaFace. If the face is not detected, the image is removed from the dataset. This is the reason of the decrease in dataset size. The cropped dataset is created by masking the areas outside the five facial landmarks. If the landmarks are not detected, the image is discarded. 

The original, aligned, and cropped datasets are merged to form a larger dataset. Although this larger dataset contains some repeated images (either original, aligned, or cropped versions), it increases the dataset size and variation by a process like offline augmentation. We call this dataset as the augmented merged dataset to avoid confusion. The data distribution of the augmented merged dataset can be seen in Figure \ref{fig:merged_dataset}. Even though the dataset is large enough in total, the minority classes disgust and fear are still underrepresented.

To avoid the overfitting of the learning algorithm to the majority classes, the following techniques have been applied: 
\begin{itemize}
    \item \textbf{Early stopping} \cite{garatti2022complexity} is used to prevent the model from overfitting by halting training when performance on the validation set stops improving.
    \item \textbf{On-the-fly data augmentation} \cite{cubuk2020randaugment} is applied during training to introduce more variations in the dataset, improving the model's robustness.
    \item \textbf{Weighted random sampling} \cite{shekelyan2022weighted}  is used to reduce the class imbalance. This technique first calculates  a weight for each class $c$ using
    \begin{equation}
        weight_c = \frac{1}{n_c/N},
    \end{equation}
    where $n_c$ is the number of samples in class $c$ and $N$ is the total number of samples in the dataset. The method then assigns different probabilities to each sample during batch sampling based on class weights $weight_c$. The high weight assigned to minority classes ensures that these images are selected more frequently during training, helping balance the dataset and improve the learning of underrepresented classes.
\end{itemize}

\subsection {Feature Learning}
\label{sec:ResNet}

In this study, we propose to use powerful CNN architectures for FER which allow deeper models while overcoming the vanishing gradient problem. Among several CNN architectures, we have used DenseNet \cite{patwal2022facial} , EfficientNet \cite{utami2022efficientnet} and ResNet \cite{zhong2017deep, kirbiz2024facial}. 

DenseNet's dense connections allow the feature values to propagate more effectively and enable deeper networks with improved performance on complex facial features \cite{patwal2022facial}. EfficientNet \cite{utami2022efficientnet} uses a compound scaling method to optimize depth, width, and resolution, ensuring high accuracy and computational efficiency. ResNet \cite{zhong2017deep}, with its residual connections, addresses the vanishing gradient problem directly which enables to train very deep networks without losing valuable information during backpropagation. Using these architectures, we can build deeper CNN models capable of capturing more complex patterns in facial expressions, which is crucial for the accuracy of FER systems.

\section{Experimental Results}
\label{sec:results}
In this section, we evaluate the performance of the proposed method. We first introduce the datasets for this experiment, followed by the explanation of the  evaluation criteria and experimental setup. Finally, the evaluation results are reported.

\subsection{Dataset}
\label{sec:dataset}

We merged three distinct datasets, FER+ \cite{barsoum2016training}, CK+ \cite{lucey2010extended}, and KDEF \cite{calvo2008facial} to evaluate the proposed methods. Table \ref{tab:compare_dataset} presents the details of the three datasets in terms of number of channels, image size, number of images, and number of emotions. The proposed FER approach recognizes seven main emotions: Angry, Disgust, Fear, Happy, Neutral, Sad and Surprise. 

The FER+ dataset originally contains images from 8 emotion classes, the 7 main emotions and an additional contempt class. The contempt class was removed from the dataset for data consistency. The images in FER+ and CK+ are grayscale images of size $48\times 48$. The images in the KDEF database are colored images of size $562\times 762$. 

\begin{table}[]
    \centering 
    \begin{tabular}{|l|c|c|c|}
    \hline 
       \textbf {Properties} &  FER+ & CK+ & KDEF\\ \hline
       \# of channels & 1 & 1 & 3 \\ \hline
       Image size & $48 \times 48$ & $48 \times 48$ & $562 \times 762$ \\ \hline
       Total images & 35,269 & 927 & 2,938 \\ \hline
       \# of classes & 8 & 7 & 7 \\ \hline
    \end{tabular}
    \caption{Comparison of the dataset properties}
    \label{tab:compare_dataset}
\end{table}

\begin{table}[]
    \centering 
    \begin{tabular}{|l|c|c|c|c|}
    \hline 
       \textbf{Emotion} &  FER+ & CK+ & KDEF & \textbf{Total}\\ \hline
       Anger & 3,110& 135& 420 & 3,665\\ \hline
       Disgust& 248& 177& 420 & 845\\ \hline
       Fear & 819& 75& 420 & 1,314\\ \hline
       Happy & 9,355& 207& 420 & 9,982\\ \hline
       Neutral & 12,905& 0& 420 & 13,325\\ \hline
       Sad & 4,370& 84& 419 & 4,873 \\ \hline
       Surprise & 4,462& 249& 419 & 5,130\\ \hline     
       \textbf{Total} &35,269 & 927& 2,938 & 39,134\\ \hline
    \end{tabular}
    \caption{Category distribution between the datasets.}    \label{tab:distribution_dataset}
\end{table}

The distribution of each dataset is presented in Table \ref{tab:distribution_dataset}. While CK+ consists of images from 7 classes, it does not have the neutral class, instead the contempt class, which was also removed. The FER+ dataset is an imbalanced dataset, with happy and neutral classes being oversampled, whereas disgust and fear classes are undersampled.  In CK+, the fear and sad classes are undersampled, whereas the other classes have more balanced distribution. Unlike FER+ and CK+, KDEF is a balanced dataset, with each class containing 419 or 420 images. 

\begin{figure}
    \begin{subfigure}   {\textwidth}        \includegraphics[width=\linewidth]{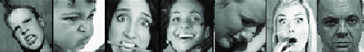}
    \label{fig:fer_plus}
    \end{subfigure}
    \begin{subfigure} {\textwidth}
   \includegraphics[width=\linewidth]{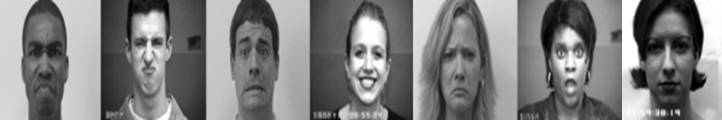}
    \label{fig:ck_plus}     
    \end{subfigure}
    \begin{subfigure} {\textwidth}
   \includegraphics[width=\linewidth]{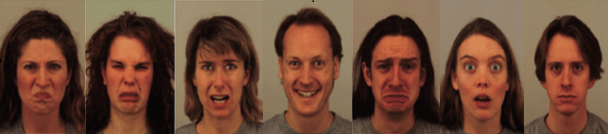}
    \label{fig:kdef}     
    \end{subfigure}   
    \caption{The images from three different datasets: FER+ (first row), CK+ (second row), KDEF (last row).}
     \label{fig:data}
\end{figure}

Figure \ref{fig:data} displays the sample images from seven basic emotions of the FER+ (first row), CK+ (second row), KDEF(last row). It can be seen from Figure \ref{fig:data}, that three datasets differ in lighting conditions, background, resolution and color space. To ensure consistency, all images are converted to gray-scale and resized to $48 \times 48$ pixels. To minimize the variations, preprocessing is applied before merging them into a single dataset. 

\begin{figure}[t]
\centering
\includegraphics[width=\linewidth]{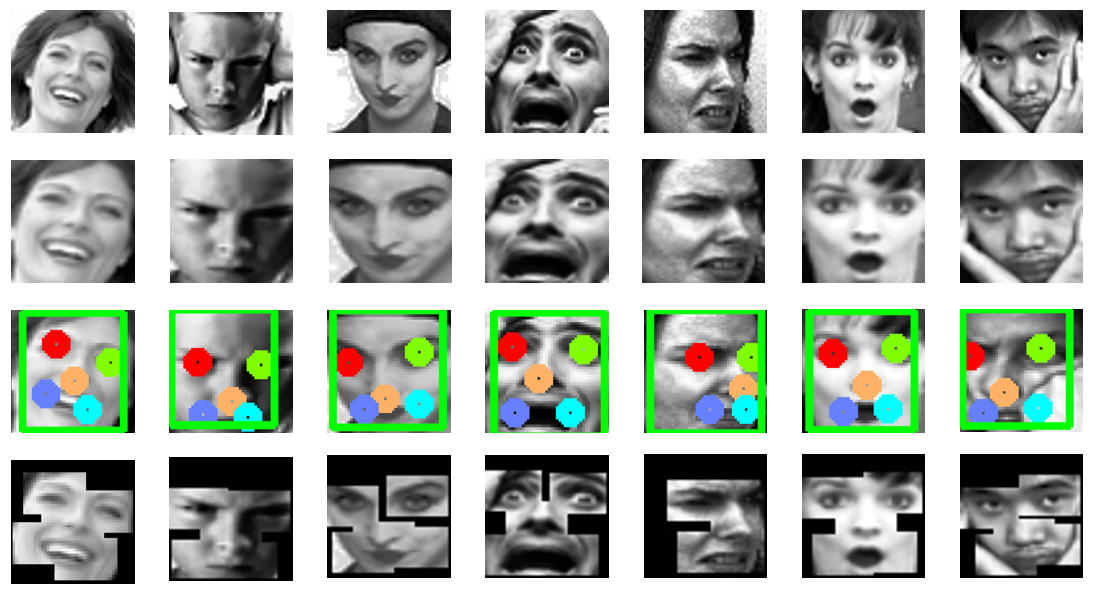}
\caption{The seven basic emotions (happy, angry, neutral, fear, disgust, surprise, sad) from the training dataset. The images in the first row are original images, the images in the second row are aligned images, the images in the thid row are the aligned and landmarked images, the images in the fourth line are face images cropped around the landmarks.}
\label{fig:database}
\end{figure}

After merging, the resulting dataset includes 39,134 facial images from the seven basic emotions. Figure  \ref{fig:database} illustrates example facial images from the merged dataset (happy, angry, neutral, fear, disgust, surprise, sad). RetinaFace \cite{deng2020retinaface} is applied to the facial images of the merged dataset to align the faces and remove the background. The second row of Figure \ref{fig:database} displays the images after alignment. These aligned faces are then resized to $48 \times 48$. In the third row of Figure \ref{fig:database}, facial landmarks are displayed together with the face boundary. The last row displays the images where a $10\times14$ rectangular mask around each landmark is applied which discards the rest of the pixels. 

The dataset is split into  $80\%$ for training, $10\%$ for validation and $10\%$  for testing. 

\subsection{Evaluation Criteria}
\label{sec:evaluation}
The performance of the proposed method is evaluated  using several metrics: precision, recall, F1 score, accuracy and confusion matrix.

Precision is defined as the ratio of the true positive predictions (TP) to all positive predictions (TP + FP):

\begin{equation}
    Precision = \frac{TP}{TP+FP}
    \label{eq:precision}
\end{equation}

Recall is the proportion of true positive predictions (TP) relative to the sum of true positive predictions (TP) and false negative predictions (FN):

\begin{equation}
    Recall = \frac{TP}{TP+FN}
    \label{eq:recall}
\end{equation}

F1 score is computed as the harmonic mean of recall and precision:

\begin{equation}
    F1Score = \frac{2\times Recall \times Precision}{Recall + Precision}
    \label{eq:F1Score}
\end{equation}

The confusion matrix is used to demonstrate the number of correctly and incorrectly predicted samples by a classifier. It is calculated based on four items: TP, TN, FP, FN. 

Finally, the accuracy rate represents the proportion of correct predictions made on the given test set.

\subsection{Experimental Details}
All experiments are conducted for 100 epochs, optimizing the cross entropy loss function. The image dataset is shuffled, and samples are selected randomly and entered into the training, validation, and test sets to ensure the reproducibility of the experiment. The Adam optimizer is employed with a learning rate of 0.001. To prevent overfitting, the model parameters are saved for the best validation accuracy. The experiments are performed on an NVIDIA Tesla K80 GPU and implemented with PyTorch.

\subsection{Results}
We report the classification performance of the proposed method using state-of-the-art CNN architectures: ResNet, DenseNet and EfficientNet. Sprecifically, ResNet models with depths 18, 34 and 50  have been used to investigate the effect of network depth on classification accuracy. 

\begin{table}
\centering%
\begin{tabular}{|l|| c | c | c|}%
\hline  
Method & Face Alignment & Landmarks & Accuracy Rate (\%)  \\ 
\hline  
DenseNet & \multirow{5}{*}{\XSolid} & \multirow{5}{*} {\XSolid} &  77.07\\
EfficientNet &  &  & 68.64\\
ResNet18 & &  & 73.81\\
ResNet34  &  &  &   73.38\\
ResNet50 &  &   & 74.45\\
\hline 
DenseNet &  \multirow{5}{*} {\Checkmark} & \multirow{5}{*} {\XSolid} & 78.49\\
EfficientNet &   &  & 73.01 \\
ResNet18  &  &  &   76.87\\
ResNet34  &  &  &  75.77 \\
ResNet50 &  &  & 75.95 \\
\hline
DenseNet &  \multirow{5}{*} {\Checkmark} &  \multirow{5}{*} {\Checkmark} & 76.33\\
EfficientNet &   &   & 68.94 \\
ResNet18  &  &  & 73.42 \\
ResNet34  &  &  &  74.76\\
ResNet50 &  &  & 74.68\\
\hline
\end{tabular}
\caption{Classification Accuracies on Test Set of the merged dataset}
\label{tab:accuracy_preprocess}
\end{table}

Table \ref{tab:accuracy_preprocess} lists the accuracy rate of the FER evaluated on the merged dataset. The first five rows reports the accuracy obtained by EfficientNetb0, DenseNet121, and ResNet models (ResNet18, ResNet34, and ResNet50).  When face alignment is applied using RetinaFace \cite{deng2020retinaface} to all data, the classification accuracy has increased by $1-5 \%$. When the aligned faces are masked around the facial landmarks, the classification accuracy increased by less $1 \%$. We can conclude that, the best FER performance can be obtained on the aligned faces. The accuracy rate is similar for the three different depths of ResNet. Thus, we will conduct the remaining tests of ResNet only using ResNet18. 

In order to see the effect of data augmentation and random weighted sampling, we have performed the tests using the ResNet18, Densenet and EfficientNet models. In \cite{kirbiz2024facial}, we have compared various agmentation methods and observed that RandAugment \cite{cubuk2020randaugment} increases the generalization and outperforms the other augmentation techniques. Thus, we have only applied RandAugment for on-the-fly augmentation. We have also applied random weighted sampling to overcome the imbalance data problem. Table \ref{tab:accuracy_aug} reports the classification accuracy obtained by ResNet18, DenseNet and EfficientNet with and without random weighted sampling and augmentation. We have seen that applying only random weighted sampling without augmentation does not increase the overall classification accuracy.  If only augmentation is applied, the classification increases by around $2-4 \%$. If both augmentation and random weighted sampling is applied during training, the accuracy increases by less than $2\%$. 

\begin{table}[t]%
\centering
\begin{tabular}{|l|| c | c | c|}%
\hline  
Method & Rnd. Weight. Samp.  & Augmentation  & Accuracy  (\%)  \\ 
\hline
\hline  
ResNet18  & \multirow{3}{*} \XSolid & \multirow{3}{*} \XSolid & 76.87  \\
EfficientNet  &  &  &  73.01 \\
DenseNet  &  &  &   78.49\\
\hline
ResNet18  & \multirow{3}{*} \XSolid & \multirow{3}{*} \Checkmark  &  80.85 \\
EfficientNet  &  &   & 75.39 \\
DenseNet  &  &  & 81.35 \\
\hline
ResNet18  & \multirow{3}{*} \Checkmark & \multirow{3}{*} \XSolid &   74.99\\
EfficientNet  &  &  &  72.52 \\
DenseNet  &  &  &   78.83\\
\hline
ResNet18  & \multirow{3}{*} \Checkmark & \multirow{3}{*} \Checkmark &   78.69 \\
EfficientNet  &  &  &  75.75 \\
DenseNet  &  &  & 80.85  \\
\hline
\end{tabular}
\caption{Classification Accuracies on the face aligned merged dataset using weighted random sampling and augmentation }
\label{tab:accuracy_aug}
\end{table}

In order to see the effect of cross-dataset performance, we conducted several experiments. We first trained our models on the training datasets of FER+, CK+, KDEF and the merged dataset separately. Then, using the obtained four models, we evaluated the test accuracy on the test set of FER+, CK+, KDEF and the merged set. The classification accuracies obtained on the four different training sets and four different test sets are reported in Table \ref{tab:cross_dataset}. We can see that, merging the dataset enhances the classification accuracies obtained on CK+ and KDEF databases. The classification accuracy obtained on the FER+ dataset has not changed. 

\begin{table}[t]%
\centering
\begin{tabular}{|l|| c | c | c| c|}%
\hline
  &  \multicolumn{4}{|c|} {Testset}  \\ \hline
Trainset  & FER+ & CK+ & KDEF & Merged \\ \hline
FER+ & 82.46 & 60.61 & 48.31 & 79.03\\ \hline
CK+ & 23.84 & 64.65 & 27.70& 25.29\\ \hline
KDEF & 33.47 & 59.60 & 83.78 & 38.37\\ \hline
Merged & 82.43 & 66.67 & 87.5 & 82.05\\ \hline
\end{tabular}
\caption{Performance obtained on single dataset and cross-dataset experiments using DenseNet and Random Augmentation}
\label{tab:cross_dataset}
\end{table}

To further increase the dataset size, we merged the original, aligned and cropped images of the FER+, CK+ and KDEF datasets, leveraging offline augmentation.  The distribution of the augmented merged dataset has been displayed in Figure \ref{fig:merged_dataset}. The classification accuracies obtained on the test set of the aligned facial images has been reported in Table \ref{tab:accuracy_merge}, where we compare the classification accuracies of ResNet18, DenseNet and EfficientNet with and without random weighted sampling and augmentation. The merging of the datasets for offline augmentation increases accuracy by $2 - 8\%$ in all test cases. Random weighted sampling does not improve overall accuracy, but applying random augmentation during training increases accuracy by $1 – 4\%$ as shown in Tables \ref{tab:accuracy_aug} and \ref{tab:accuracy_merge}.    

\begin{figure}
    \begin{subfigure}   {0.55\textwidth}        \includegraphics[width=\linewidth]{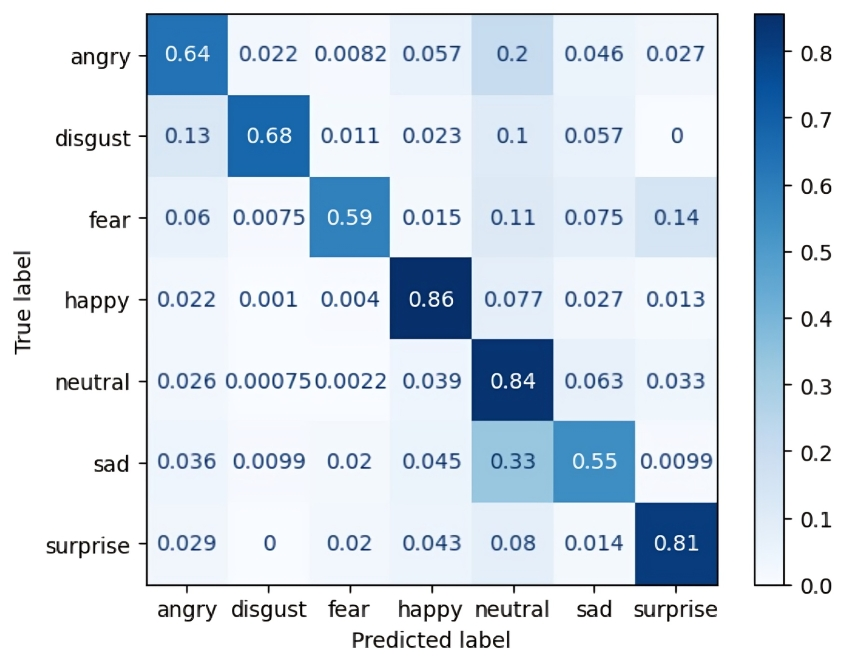} 
    \caption{Baseline tested on merged dataset}
    \label{fig:confusion_base}
    \end{subfigure}
    \begin{subfigure} {0.55\textwidth}
   \includegraphics[width=\linewidth]{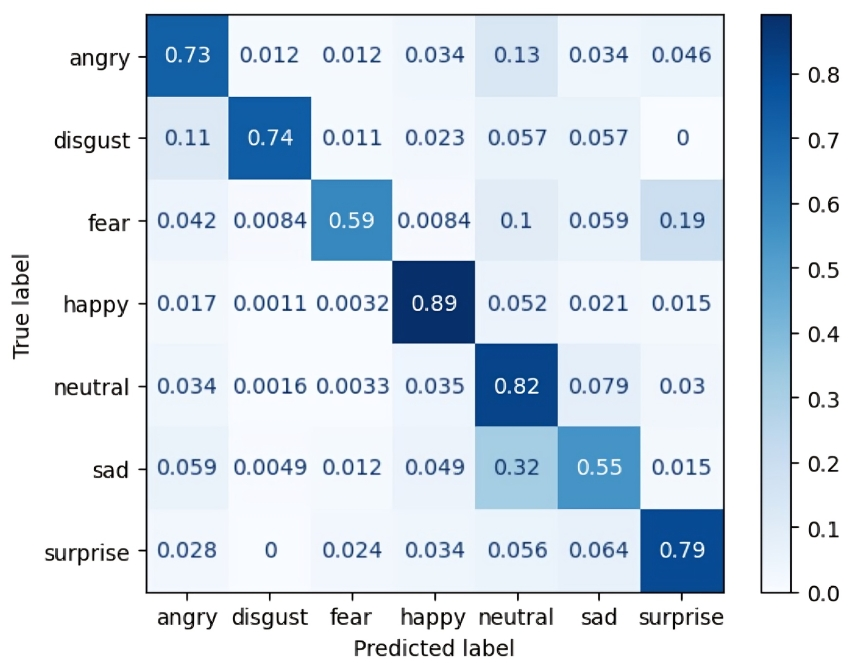}
    \caption{Baseline tested on aligned merged dataset}
    \label{fig:confusion_aligned}     
    \end{subfigure}
    \begin{subfigure} {0.55\textwidth}
   \includegraphics[width=\linewidth]{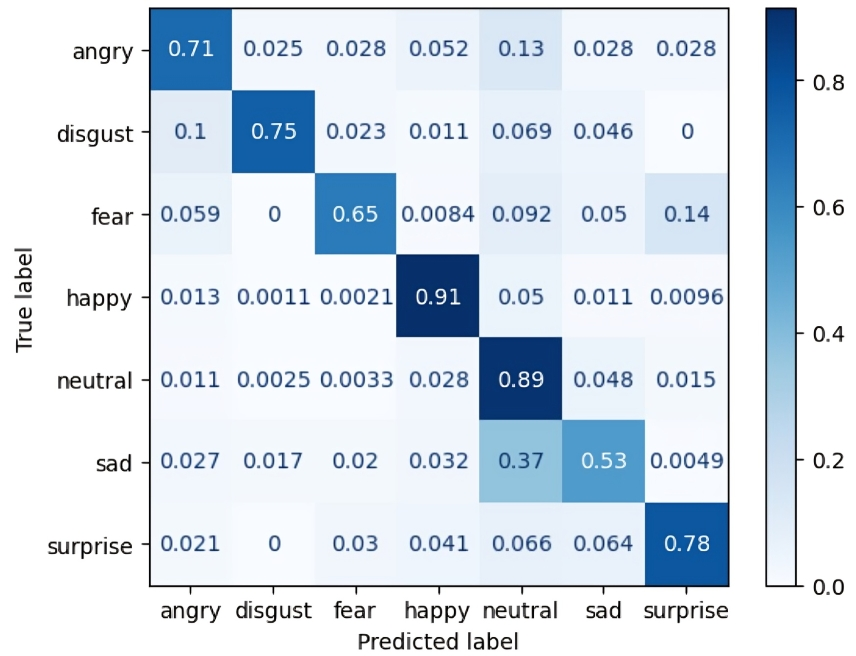}
    \caption{RandAug tested on aligned merged dataset}
    \label{fig:confusion_randaug}     
    \end{subfigure}   
    \begin{subfigure} {0.55\textwidth}
   \includegraphics[width=\linewidth]{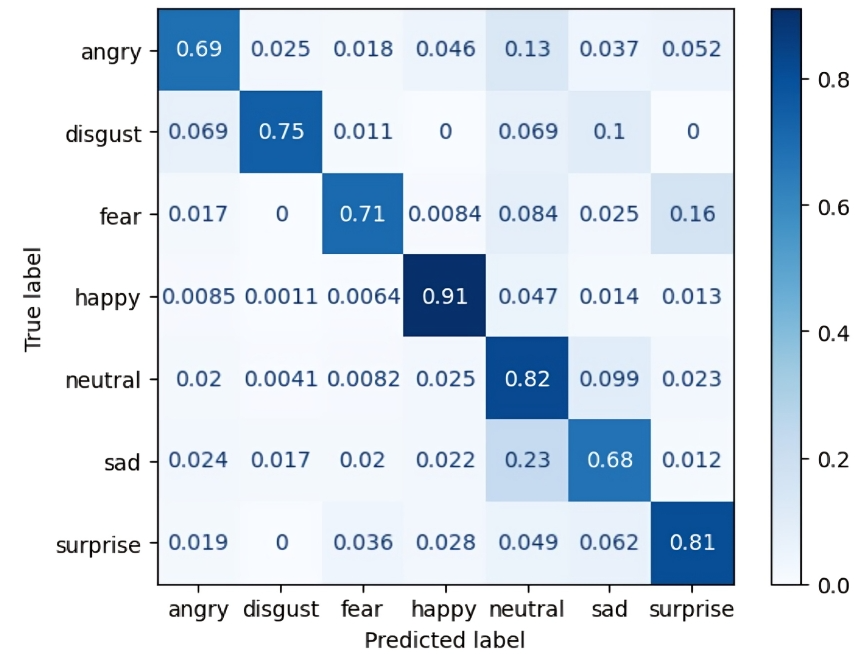}
    \caption{RandAug and RandomWeighted Sampler tested on aligned merged dataset}
    \label{fig:confusion_randaug_weightedsampler}         \end{subfigure}      
    \begin{subfigure} {0.55\textwidth}
   \includegraphics[width=\linewidth]{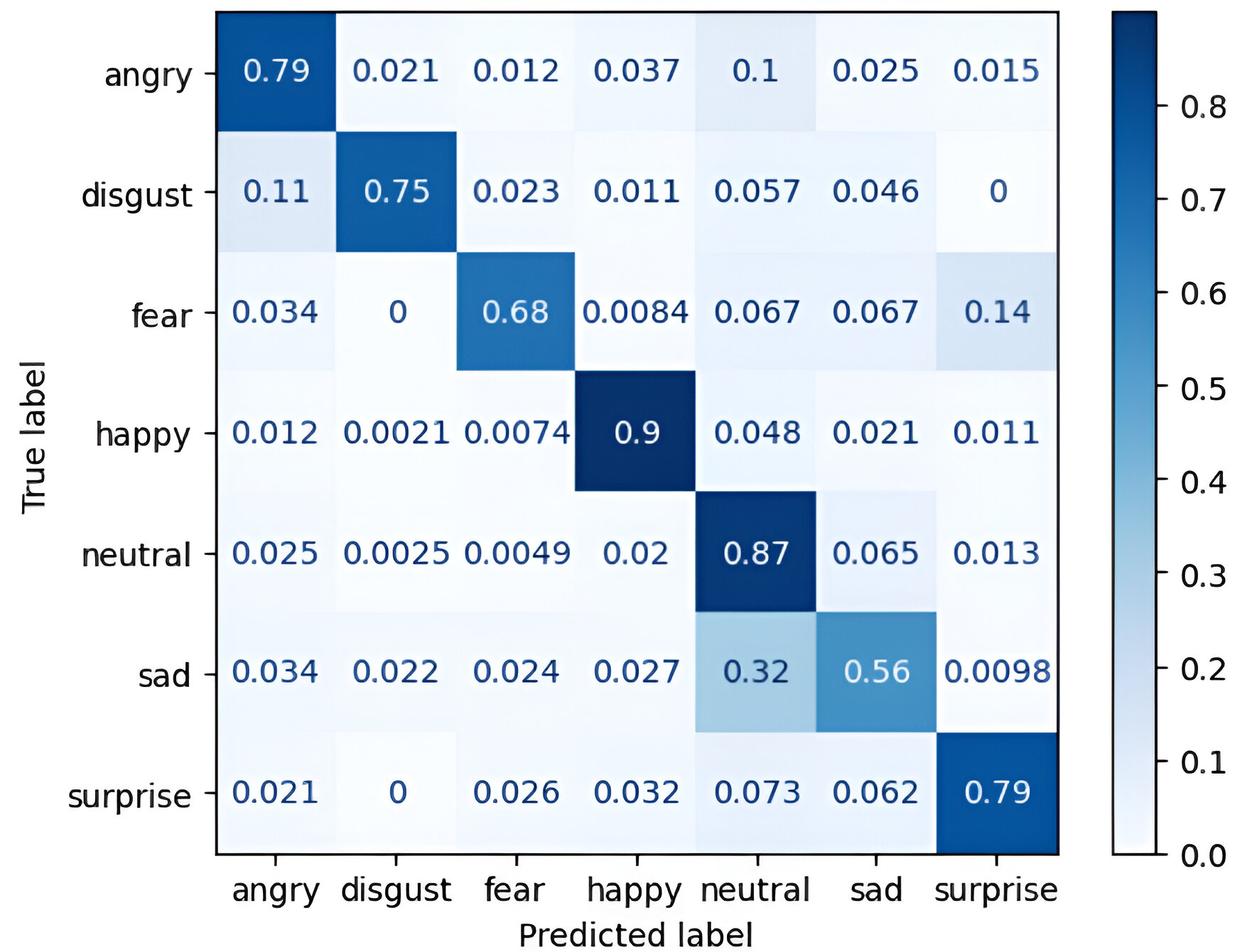}
    \caption{RandAug tested on augmented merged dataset (Original + Aligned + Cropped)}
\label{fig:confusion_merged_randaug}         \end{subfigure}      
    \caption{The confusion matrices obtained by the DenseNet121 model}
     \label{fig:confusion}
\end{figure} 

In order to see the effect of face alignment, on-the-fly augmentation, random weighted sampling and offline augmentation in terms of the classification accuracies for each class, we represent the confusion matrices in Figure \ref{fig:confusion}. Figure \ref{fig:confusion_base} represents classification accuracies obtained by the baseline DenseNet architecture on the merged dataset. We can see the effect of our contributions by comparing the confusion matrices in Figure \ref{fig:confusion_aligned}- \ref{fig:confusion_randaug_weightedsampler} with the baseline in Figure \ref{fig:confusion_base}. Face alignment enhances the classification accuracies for angry, disgust, happy, and surprise classes whereas the performance does not change for the other classes. Random Augmentation enhances the classification accuracy for the minority class fear and random weighted sampling applied together with random augmentation enhances the classification accuracy for the minority class sad.  
 
Finally, we analyze the effect of random weighted sampling on the precision, recall and F1 scores for for each emotion class and report in Table \ref{tab:precision_recall}. The tests are conducted with and withot random weighted sampling on the test set. Although random weighted sampling does not increase the overall classification accuracy, it increases the precision, recall and F1 score for the minority classes disgust and fear.  

\begin{table}[t]%
\centering
\begin{tabular}{|l|| c | c | c|}%
\hline  
Method & Rnd. Weight. Samp.  & Augmentation  & Accuracy  (\%)  \\ 
\hline
\hline  
ResNet18  & \multirow{3}{*}\XSolid & \multirow{3}{*} \XSolid &   80.85\\
EfficientNet  &  &  &  75.11 \\
DenseNet  &  &  &  81.01 \\
\hline
ResNet18  & \multirow{3}{*}\XSolid & \multirow{3}{*}\Checkmark  &  82.05 \\
EfficientNet  &  & & 77.03    \\
DenseNet  &  &  &  81.50 \\
\hline
ResNet18  & \multirow{3}{*}\Checkmark & \multirow{3}{*}\Checkmark &   80.79 \\
EfficientNet  &  &  & 76.05  \\
DenseNet  &  &  &  80.93 \\
\hline
\end{tabular}
\caption{Classification Accuracies on the face aligned merged dataset trained on the augmented merged dataset using weighted random sampling and augmentation }
\label{tab:accuracy_merge}
\end{table}

\begin{table}[t]%
\centering
\begin{tabular}{|l|| c | c | c|| c | c|c|}%
\hline  \
& \multicolumn{3}{c||} {\bf{Without Rnd Weight Samp.}}& \multicolumn{3}{c|}{ \bf{With Rnd Weight Samp.}} \\
\hline 
\bf{Emotion} & \bf{Precision}  & \bf{Recall}  & \bf{F1 Score} & \bf{Precision}  & \bf{Recall}  & \bf{F1 Score } \\ 
\hline 
Angry  &0.809& 0.758 &0.782&0.806& 0.774 &0.789    \\
\hline
Disgust  &0.800 &0.790 & 0.795&0.849 &0.816 & 0.832\\
\hline
Fear& 0.807& 0.628& 0.707& 0.845& 0.628& 0.721 \\
\hline
Happy &0.897 & 0.920& 0.908&0.892 & 0.903& 0.898 \\
\hline
Neutral &0.806 & 0.879& 0.841&0.804 & 0.862& 0.832 \\
\hline
Sad & 0.725&0.572 & 0.640& 0.677&0.594 & 0.633 \\
\hline
Surprise & 0.854& 0.837& 0.855& 0.856& 0.837& 0.846 \\
\hline
\end{tabular}
\caption{Classification performance in terms of Precision, Recall and F1 Score}
\label{tab:precision_recall}
\end{table}

\section{Conclusion}
\label{sec:conclusion}
In this work, we proposed an automatic facial emotion recognition (FER) method using deep CNN architectures DenseNet, ResNet and EfficientNet. We enhanced the training process by integrating several strategies, including data merging from three different datasets (FER+, CK+, and KDEF), face alignment using RetinaFace, and the application of both online and offline data augmentation techniques. Additionally, we addressed the class imbalance problem by incorporating random weighted sampling during training.

Our experimental results show that all the architectures benefit significantly from both online/offline data augmentation and random weighted sampling. These techniques improved classification accuracy and enhanced the model's generalization ability, especially for minority classes such as disgust and fear. The combination of augmented data and face alignment provided the best results, leading to higher classification accuracy.

In future work, we plan to extend our dataset to include more realistic and diverse real-world datasets to evaluate the model's robustness in challenging, real-world scenarios. Additionally, we aim to explore the potential of deeper network architectures and incorporate transfer learning to further improve model performance. Another important direction for future research will be integrating interpretability methods, such as Gradient-weighted Class Activation Mapping (Grad-CAM) \cite{araf2022real}  and SHAP \cite{lorch2025towards}, to gain deeper insights into the model's decision-making process and enhance its transparency. Furthermore, we recognize the dynamic and context-dependent nature of emotions, and as part of future work, we plan to explore temporal modeling techniques to capture the evolving nature of emotions in video-based or real-time scenarios. We also aim to incorporate context-aware methods to improve recognition accuracy by considering surrounding cues, such as environmental or physiological data. 







\bibliographystyle{elsarticle-num-names} 
 \bibliography{refs.bib}



\end{document}